\documentclass[sigconf,nonacm]{acmart}

\AtBeginDocument{%
  \providecommand\BibTeX{{%
    \normalfont B\kern-0.5em{\scshape i\kern-0.25em b}\kern-0.8em\TeX}}}

\usepackage{subcaption}
\usepackage{multirow}
\usepackage{natbib}
\setcitestyle{square, comma, numbers,sort&compress, super}

\begin{document}

\title{Panel Transitions for Genre Analysis in Visual Narratives}

\author{Yi-Chun Chen}
\affiliation{%
  \institution{Computer Science, North Carolina State University}
  \country{USA}
  }
\email{ychen74@ncsu.edu}

\author{Arnav Jhala}
\affiliation{%
  \institution{Computer Science, North Carolina State University}
  \country{USA}
  }
\email{ahjhala@ncsu.edu}

\renewcommand{\shortauthors}{Chen et al.}

\begin{abstract}
Understanding how humans communicate and perceive narratives is important for media technology research and development. This is particularly important in current times when there are tools and algorithms that are easily available for amateur users to create high-quality content. Narrative media develops over time a set of recognizable patterns of features across similar artifacts. Genre is one such grouping of artifacts for narrative media with similar patterns, tropes, and story structures. While much work has been done on genre-based classifications in text and video, we present a novel approach to do a multi-modal analysis of genre based on comics and manga-style visual narratives. We present a systematic feature analysis of an annotated dataset that includes a variety of western and eastern visual books with annotations for high-level narrative patterns. We then present a detailed analysis of the contributions of high-level features to genre classification for this medium. We highlight some of the limitations and challenges of our existing computational approaches in modeling subjective labels. Our contributions to the community are: a dataset of annotated manga books, a multi-modal analysis of visual panels and text in a constrained and popular medium through high-level features, and a systematic process for incorporating subjective narrative patterns in computational models. 
\end{abstract}

\maketitle

\section{introduction}


Understanding how humans create and process multi-media narratives is important in many fields ranging from media studies, engineering, cultural studies, and media production. Many techniques have been developed over the years in the multimedia community to process and understand textual, visual, and auditory media. In particular, a strand of research called computational media aesthetics looked at analysis of multi-media artifacts to understand narrative phenomena such as pacing/tempo, film shot transitions, etc. Prior work in this area and in the are of signal processing level multimedia analysis has provided us with tools to now start connecting low-level features to higher-level semantics. One challenge in systematically addressing this topic is the availability of data that represents how media practitioners apply creation patterns of narrative to make them appeal to a wide audience. While in computer vision and natural language processing, there are several datasets that come with annotations for training models for tasks like object detection or sentiment analysis, these are still at a much lower-level of perception in terms of storytelling.

In this paper, we address the larger challenge of connecting low-level media features to higher-level narrative semantics. In particular, we present an annotation scheme for panel transitions in manga style visual narratives. The annotations are chosen from widely known best-practices in comic creation that are also utilized in the development of cognitive reading models of narrative comprehension. We introduce annotations of panel transitions in the Manga109 dataset (we will release the data and associated annotations with the publication of this paper). We provide a systematic statistical analysis of the annotation process including a characterization of the subjective aspects of annotation across human annotators. To provide a concrete practical example of the use of this dataset, we address a novel problem of computational analysis and classification of {\em genre} in manga. We show with an ablation study how the performance of an LSTM-based model improves in the task of detecting genre through the introduction of panel transition features. We also present deeper analysis of the patterns of panel transitions that are observed in manga books through Generalized Sequence Mining algorithms. Overall, we conclude that providing the community with a novel annotated dataset, connecting annotations with practitioners authoring process, and providing methods for rich narrative-level analysis of media will improve our understanding of multi-media artifacts and authoring process.

We chose manga as our media form for three reasons. First, it is rich in narrative as well as stylistic content with both text and visual modalities. Second, it is easier to process compared to videos that have a lot of redundancy between frames. In manga and comics each panel is carefully composed to be as efficient with the presentation of information as possible. Third, it has established cultural and artistic traditions that differ from western comics, thus allowing future research from several different perspectives of storytelling. 

The rest of the paper is structured as follows: We begin in Section 2 with related work in the area of genre detection for different media forms such as text and movies. We then introduce previous work specifically in the analysis of comics and manga style narratives. In section 3, we describe our annotation scheme and provide information about the dataset and our labeling process. This is described in detail due to the potential subjectivity in the labeling task. Finally in Section 4 we describe experiments for genre detection with raw features and with labels. We also analyze panel transitions based on their usage in books of different genres that were extracted with the genre detection algorithm. 

\section{related work}

Genre study based on content is a well established research area within multimedia analysis, especially in terms of recommender systems and for building knowledge graphs for information retrieval. It has been studied separately across multiple media forms including audio\cite{simeone2017hierarchical}, textual narratives\cite{tafreshi2018emotion}, web documents \cite{pritsos2019open}, films \cite{choros2018video}\cite{doshi2018movie}\cite{shambharkar2020survey}\cite{yadav2020unified}, and comics\cite{daiku2018comics}\cite{park2019estimating}. 
Among these film and comics are forms that combine both visual and textual/audio media. 

In the movie area, genre identification has seen significant research by using different components for indexing, including generation of trailers \cite{shambharkar2020survey}\cite{yadav2020unified}, posters \cite{kundalia2020multi}, topological metadata \cite{doshi2018movie}, frame content \cite{sreeja2019towards}, and the temporal structure\cite{yu2020asts}\cite{choros2018video}. Analyses substantiated the theory that genres highly influence film structures; moreover, previous research showed that the structures helped retrieve genres. Although Comics, as a medium, is older than film and share similar sequential narrative characteristics\cite{pratt2009narrative}, the genre detection in this area fails to have a similarly broad discussion across different aspects other than the image content.

In previous research, the comic genres were used to summarize story content that potentially serves the query based on the reader’s interest \cite{daiku2017comic}. Furthermore, to better represent the story detail, the image contents of pages in a comic book were described by sub-sequence that consist of genres \cite{doshi2018movie}.  While the genres encompass the image contents, the discussions did not consider the contiguous panels and the relationships between panels. Those narrative aspect features are usually rather neglected in comic researches, even if they are the nature of comics.

To bridge the gap between the image content, the usual emphasis of computer vision, and comics' narrative nature. Iyyer et al. used hierarchical LSTM to capture the image content sequentially and analyzed the "gutters" between panels based on McCloud's definition of panel transitions \cite{iyyer2017amazing}. Yet, the link between narrative features, derived in sequential content, was not discussed further. The theory of the narrative aspects features benefit sequential content understanding were demonstrate in previous researches. McCloud’s books about comic understanding proposed transitions that cover the time and space shifts between panels \cite{mccloud1998understanding}\cite{mccloud1993understanding}\cite{mccloud2006making}. Martens et al. then discussed the narrative and event structure helped on comic sensemaking \cite{martens2020visual}. These researches inspired our questions.


In film, work by Choroś {\em et al}~\cite{choros2018video} shows that structural analysis of videos can be useful in identification of genre. We take inspiration from this work and demonstrate that structural analysis of manga style can help with genre detection. In Cognitive Science, the Parallel Inference of Narrative Semantics (PINS) theory~\cite{cohn2019pins} proposes a structure of visual discourse presented in comics based on the role of panels in comprehension of the overall narrative. The Visual Narrative Grammar (VNG) represents the narrative structure level in the PINS model, it argues that the combinatorial structure functions to organize the meaningful information into comprehensible sequences. The VNG operates on sequential images as the syntactic structure in sentences. It gives image units a categorical roles like syntax decomposes sentences into nouns, verbs and so on. And then it organizes the categorical roles through a constituent structure to form the sequential images. This theory brings a more formal linguistic vocabulary to constituent elements of visual stories.

\section{methods}
We are trying to test the hypothesis that if panel transitions are intentionally chosen to indicate particular types of narrative events then a computational model will be able to learn these patterns of transitions. Further, we demonstrate the applicability of this result to show that genre, defined as a collection of artifacts with similar narrative arcs, can be better detected with the knowledge of transitions labels given to the system. To answer this question, we employed a corpus of visual narratives in the Manga109 dataset and then enhanced the dataset by adding panel transition narrative labels. After that, the labeled features were used to retrieve clusters. Finally, the results were compared with the baseline provided in the original dataset.  

\subsection{Target dataset}
Manga109 is a dataset of manga images. The manga is a specialized art-style of comics that developed and popularized in Japan. The dataset was first introduced into the Computer Vision field by Matsui et al. in 2017 \cite{daiku2017comic}. The dataset consists of 109 manga titles drawn by professional manga authors. Each title includes 1964 pages on average, with a total of 21,142 pages. We choose this dataset as our target because they are from real mangas that previously published in manga magazines and cover various kinds of genres: humor, battle, romantic comedy, animal, science fiction, sports, historical drama, fantasy, love romance, suspense, horror, and four frame cartoons.  It is clearly suitable for studying the links between narrative features and genres. 

\subsection{Labeling for Narrative Transitions}

To capture the medium's narrative aspects, we then analyzed different factors according to comic theories concluded by Scott McCloud \cite{mccloud1993understanding} \cite{mccloud1998understanding}, which are usually underlaid inside comics. We then first took the sequence properties shared by the films and comics into consideration. Movies using camera shots to link frames and guiding readers by switching focus between content; an analog idea in comics is the transitions between comic panels. The transitions lead readers' minds by shifting between the content too. Comic panels fracture both time and space, offering a jagged staccato rhythm of unconnected moments like single frames. It is the audience's imagination to make the "gutters" closure; therefore, authors deliberately cooperate various transitions to convey their story in a rhythm that matches their mind and emphasis the content they want to raise readers' attention. The examples of the transitions and complex ones are in figure \ref{fig:multiple_transition_pairs}.

\subsubsection{Comic transitions}

\begin{figure}
\centering
    \begin{subfigure}{.25\textwidth}
        \centering
        \includegraphics[width=0.8\linewidth]{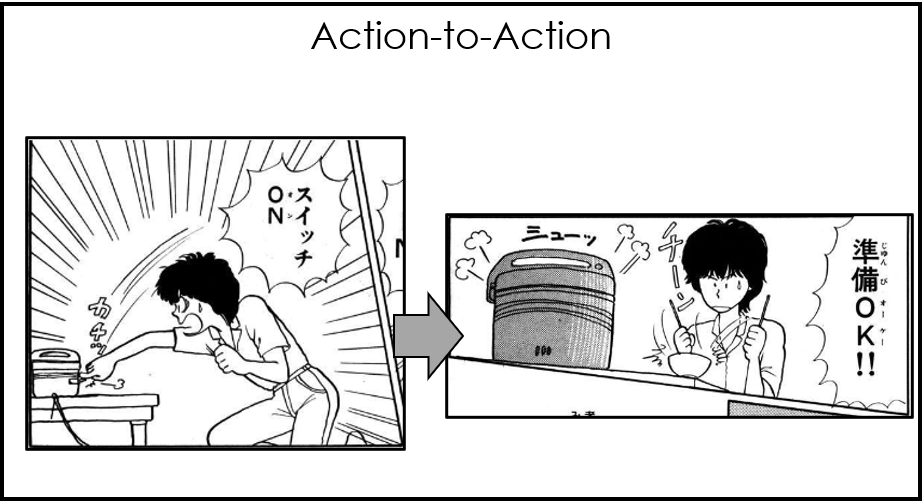}
        \caption{The scene and subject remain the same, but the character's action changed.}
        \label{fig:action_transition_example}
    \end{subfigure}%
    \begin{subfigure}{.25\textwidth}
        \centering
        \includegraphics[width=0.8\linewidth]{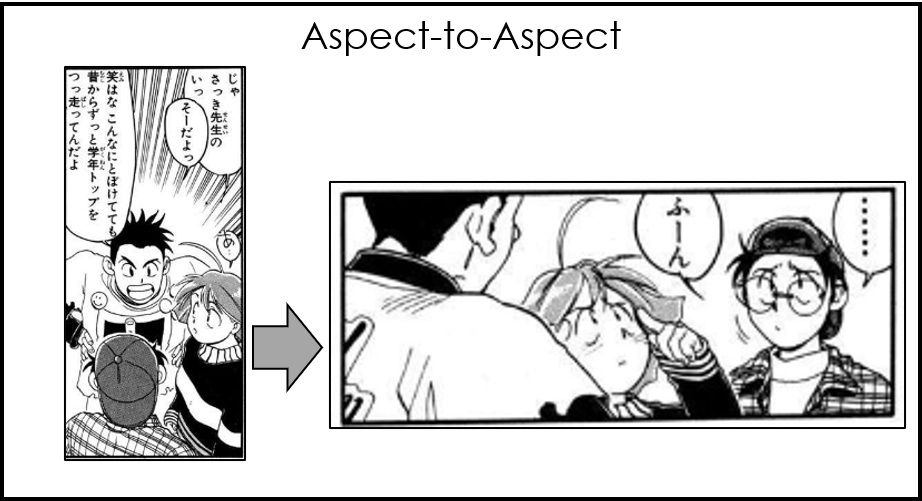}
        \caption{The subjects in the image are the same as the previous panel, but the view angle changes to emphasize the characters' reaction.}
        \label{fig:aspect_transition_example}
    \end{subfigure}
    \begin{subfigure}{.25\textwidth}
        \centering
        \includegraphics[width=0.8\linewidth]{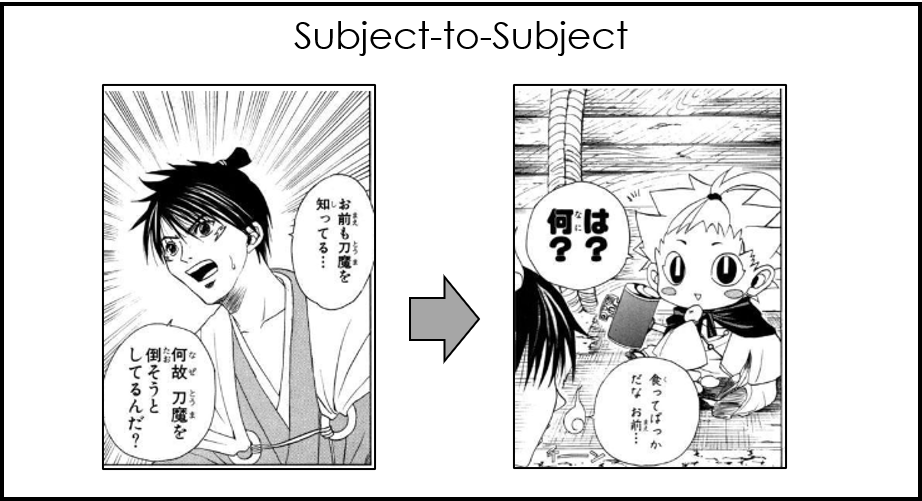}
        \caption{The focused subject changed.}
        \label{fig:subject_transition_example}
    \end{subfigure}%
    \begin{subfigure}{.25\textwidth}
        \centering
        \includegraphics[width=0.8\linewidth]{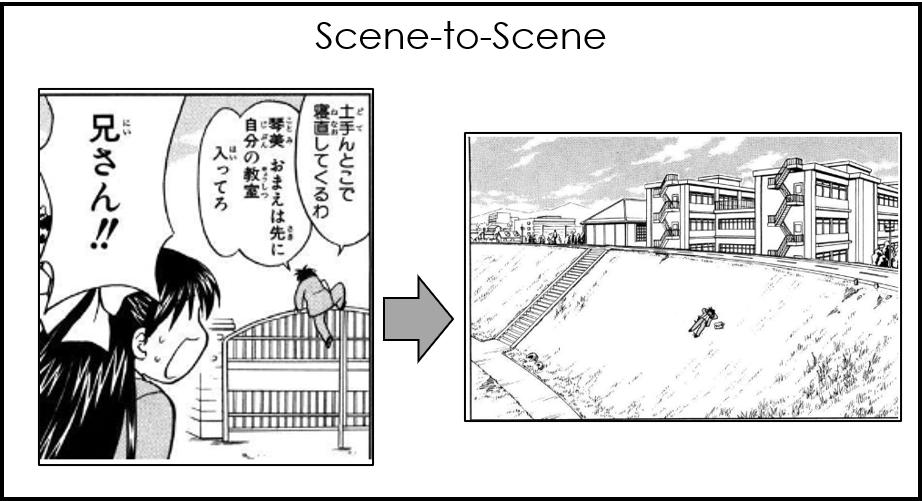}
        \caption{A significant time or space shift between panels.}
        \label{fig:scene_transition_example}
    \end{subfigure}
    \begin{subfigure}{.25\textwidth}
        \centering
        \includegraphics[width=0.8\linewidth]{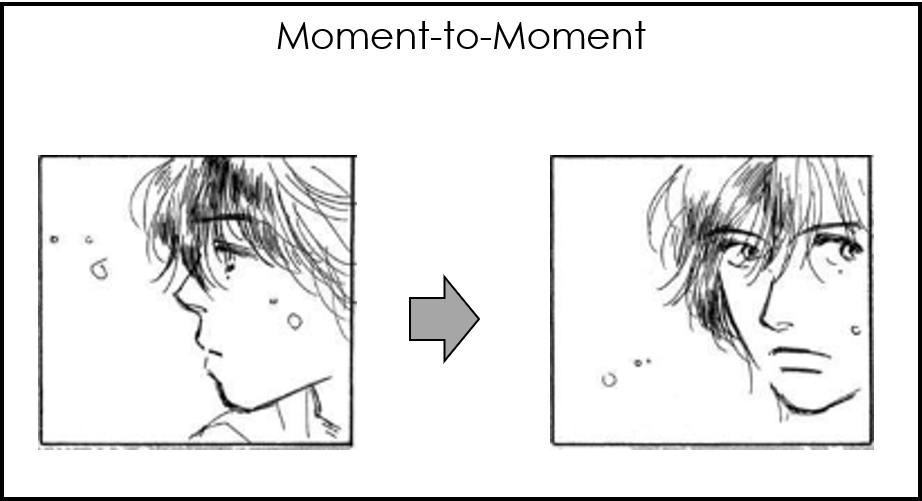}
        \caption{Between the panels, only small changes with almost no time shifting.}
        \label{fig:moment_transition_example}
    \end{subfigure}%
    \begin{subfigure}{.25\textwidth}
        \centering
        \includegraphics[width=0.8\linewidth]{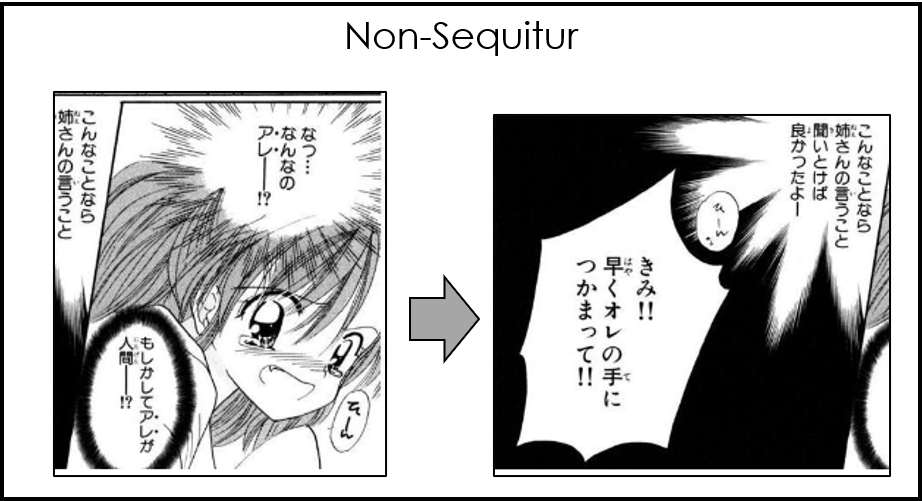}
        \caption{Focused subject shifting happened together with camera shot changes.}
        \label{fig:non_transition_example}
    \end{subfigure}    
    \caption{The transition examples corresponding to the six labels. Action to action, aspect to aspect, subject to subject, scene to scene, moment to moment, and non-sequitur.}
    \label{fig:transition_examples}
\end{figure}

There are six types of common transitions in total; sometimes, the “gutter” between the panels combines multiple transitions simultaneously to transport a rather complex idea. We manually annotated the randomly selected 2228 pairs of consecutive panels from manga pages to analyze inter-panel transitions. Each pair of consecutive panels was labeled with the transition category that best describes the change between them. Below are the definition and brief explanations of the transition types and the percentages from our annotations.The figure \ref{fig:transition_examples} gives example of each type of transitions.\\
 \\
\textbf{Action-to-action, 33.2\%}:  This transition shows a subject doing an action over a sequence of panels. In some cases, actions begin and end at different frames with a number of intermediate panels. We are only looking at pairs of panels in our annotations so we consider this annotation whenever any part (beginning, middle, end) of the same action is present in both frames. \\
\textbf{Aspect-to-aspect, 8.3\%}: This transition indicates a shift to an abstract scene similar to an interlude to indicate passing time such as a falling leaf. This is an important category for manga as it is commonly used to slow down or emphasize characters’ actions, their feelings, and moment of an event is to change the view angles of the same subject.  We categorize this kind of change as aspect transitions as well.\\ 
\textbf{Subject-to-subject, 20.4\%}: This transition indicates change or introduction of subjects between panels. In manga, there is a common method that authors often use to guide the reader’s attention, which is to change focus between a group of subjects. We also consider these changes as members of this category even if the last focused subject may still exist in the next panel. \\
\textbf{Scene-to-scene, 10.1\%}: This transition transport the reader across a significant distance of time and space between two panels.\\
\textbf{Moment-to-moment, 12.6\%}: Moment-to-moment transition captures small changes with very little to no passage of time in the story world between the two panels.\\
\textbf{Non-sequitur, 15.1\%}:   This transition indicates that there is no perceivable relationship between panels.

The annotators involved in this process were 3 students familiar with comic reading but are not professional comic producers. Different chucks and overlap groups of selected manga pairs were assigned to different annotators. The agreement among the annotators on the overlapped pairs was in figure \ref{ground truth}, and evaluated by Cohen's kappa score.

\subsubsection{Story analysis method}
The framework was presented in figure [3]. The whole dataset was divided into two sets, one is holding out and the other is training samples.  The labeled ground truth was the first set of training data, then some holding out samples were feed into model to obtain the feedback. After that, a group of unlabeled images was mixed into the framework, meanwhile, the combination of labeled data pool and the hold out set changed to obtain new feedback. Then the training process was launched again with the enlarged labeled pool.
The model we used here was transfer learning based on CNN. On top of the model, two regular deeply connected neural network layers were added to transfer the layer output from feature descriptors to 256 units and then defined the output in the next layer to match the multiple classes we wanted. The images were described by features obtained from the fc7 layer of pre-trained ImageNet weights. The implementation details including loss function, activation functions in layers, and the optimizer will be discussed in experiment. 

To provide the model sufficient information to identify the transitions, the images were considered in pairs. The input data combined two consecutive panels and the inter-panel transition labels.

\begin{figure}[ht]
    \centering
    \includegraphics[width=8cm]{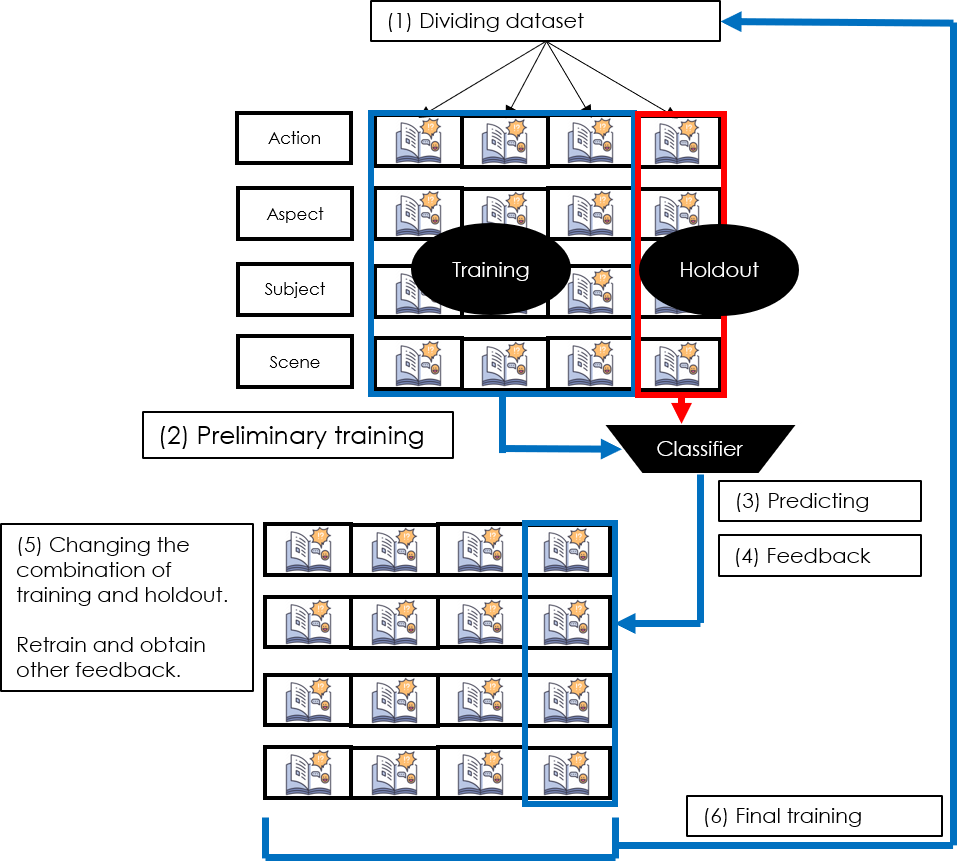}
    \caption{The labeling framework}
    \label{fig:labeling framwork that shows how the dataset is organized across the steps of the iterative refinement process of labeling.}
\end{figure}

\subsubsection{Narrative structures}
After obtained transition labels, our next step was to retrieve the narrative sequence from the results. Our transition analysis has two scopes: one was the summary of transitions that consist of a book. According to comic theories, the inter-panel transitions compressed space and time shifts, which connect the fractured parts into a complete storyline. Therefore, it is an analog concept that captures the temporal shifting relation. Our intuition here was that: like in cinema have different narrative pace in different movie genre, comics were likely to tend to use various transitions to convey the content to fit story pace. We showed that clusters of books could be coordinated by the distribution of transitions in the experiment section.

The other scope of narrative analysis was page-wise sequences. The same comic pages' transitions were extracted and formed a single sequence that represents the page’s narrative structure. We then interpret the narrative structures from two reverse sides. We first do the clustering again to see whether the sequences could show some tendency and form different groups—the next analysis laid on the genres from Manga109. We picked representative comics in different genres to see what sequences were used to transport the stories.
\subsection{Clustering}
The clustering method we used in this paper was K-means. The number of centroids was decided by comparing the distortion and inertia through the elbow method.   After the clustering process finished, we then colored the books depends on their cluster labels and then compare the intersections between the cluster and the real genre groups to see whether the clustering results overlap with genres. A detailed description of experiment settings and results was in the experiment section. We choose an unsupervised clustering method here because we suspect that the content of a book might be a combination of genres and could not be described by only one word. Although the chosen label may close the main theme of a story, the content would likely be much complicated than the extent that a single word can describe. The previous comic genre research \cite{daiku2018comics} also pointed out this similar thought. That is why we would like to use unsupervised clustering to approach the closest genres rather than direct label books with genres. This can describe the relatedness of books and can also see whether book genres have a correlation with the transition analysis through their overlapping.




\section{experimental results}
This section will present our results with the parameter choices and then show whether the use of transitions related to comic genres. 

We present our results through the following structure. First, we will analyze the ground truth labels used in our training. Then, the next sub-section evaluates the performance of the labeling framework and how well the automatically labeled results match with annotators’ feedback. After that, the learned data are discussed by two scopes, through clustering and sequence analysis. Finally, we compare our results and the genres of Manga 109.

\subsection{Ground truth analysis}

We have annotated 2228 pairs of panels as the ground truth about inter-panel transitions. In the annotating process, randomly selected pairs were dividing into an evaluation set (129 pairs) and a test set.   Each annotator got both the whole evaluation set and part of the test set. Table \ref{ground truth} shows the statistic and agreement on the overlapped evaluation set. The agreements were evaluated through cohen's kappa \cite{cohen1960coefficient}. It is a statistic that is used to measure inter-rater reliability categorical data. Cohen suggested the Kappa result be interpreted as follows: values $\leq$ 0 indicating no agreement and 0.01–0.20 as none to slight, 0.21–0.40 as fair, 0.41– 0.60 as moderate, 0.61–0.80 as substantial, and 0.81–1.00 as almost perfect agreement \cite{mchugh2012interrater}.

\begin{table*}
\centering
\begin{tabular}{|p{3.5cm}|p{2cm}|p{2cm}|p{2cm}|}
    \hline
    -- & Annotator 1$\&$2 & Annotator 2$\&$3 & Annotator 1$\&$3\\
    \hline

    \hline
    reliability (Kappa score) & 0.524 & 0.631 & 0.774\\
    \hline    
\end{tabular}
    \caption{
    \label{ground truth} \small{The agreement among annotators on evaluation set data}
    }
\end{table*}

\begin{figure}
\centering
    \begin{subfigure}{.25\textwidth}
        \centering
        \includegraphics[width=0.8\linewidth]{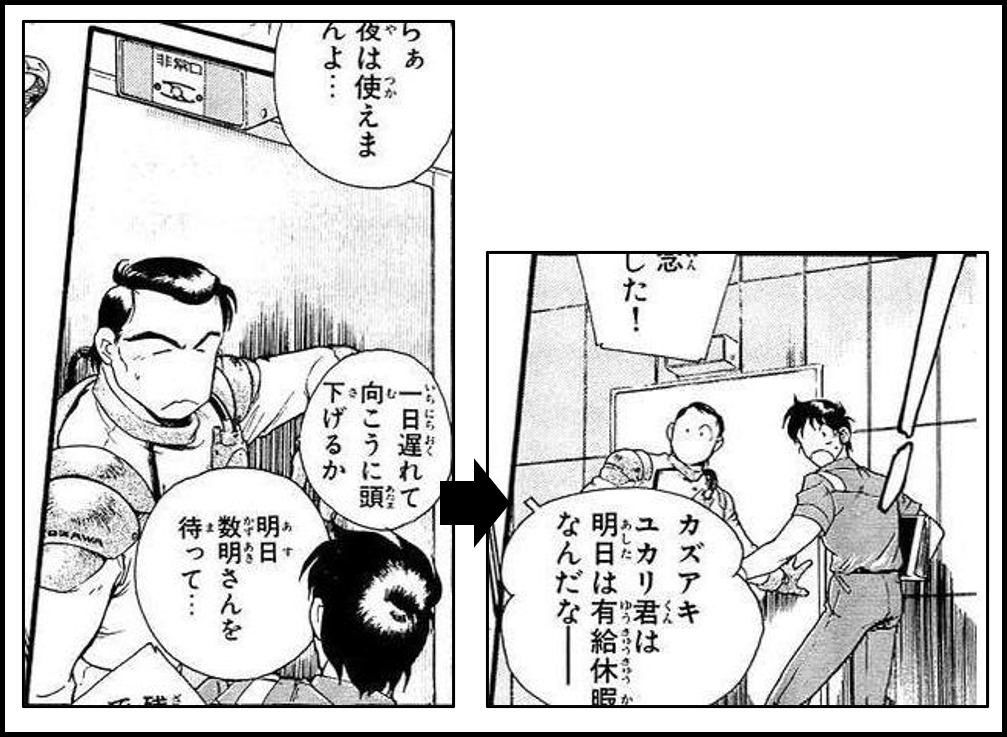}
        \caption{View angle changes happened together with action changes.}
        \label{fig:action_aspect_example}
    \end{subfigure}%
    \begin{subfigure}{.25\textwidth}
        \centering
        \includegraphics[width=0.8\linewidth]{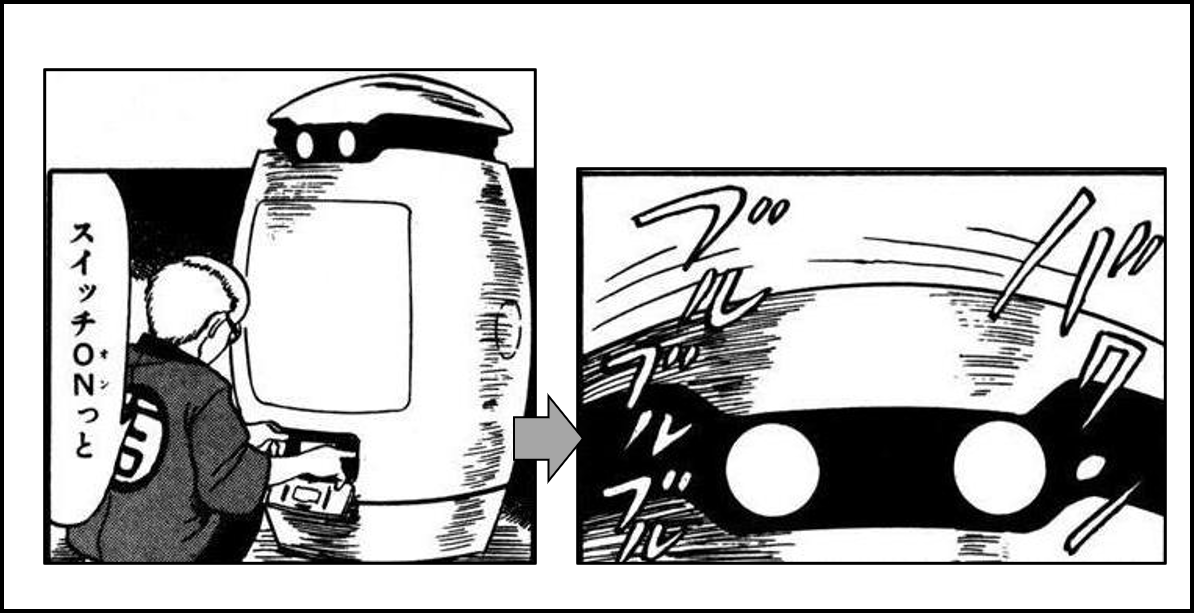}
        \caption{Focused subject shifting happened together with camera shot changes.}
        \label{fig:subject_aspects_example}
    \end{subfigure}
    \caption{Panel pairs with multiple transitions.}
    \label{fig:multiple_transition_pairs}
\end{figure}

In some cases, the transition between a panel pair is a combination of multiple types of transitions rather than one that can be easily fit into a category. The multiple possibilities of labels were the reason why annotators might disagree. The figure \ref{fig:multiple_transition_pairs} gives a few examples.  In figure \ref{fig:action_aspect_example}, the subjects of the first panel are two characters (from left to right); in the next panel, the subject remained the same, the view angle changed to capture the characters' emotional reaction. Meanwhile, their action also different from the previous panel. In this situation, the annotators predicted the author's intention and chose a label they thought to capture the situation well. Another example in figure \ref{fig:subject_aspects_example} showed that when the focused subjects in the panel changed from two characters to one character, at the same time, the view shot become a close shot to emphasize what is happening to the character.  Therefore, the transition between the panel can be interpreted as both subject-to-subject or aspect-to-aspect. The kappa scores in the table \ref{ground truth} suggested the reliability between our annotator above or close to a substantial level.

\subsection{Performance discussion on labeling model}
In order to test whether the feedback-training framework benefits the learning process, in this experiment we compare both the model learning accuracy and labeling reliability between two sets. 

\begin{figure}[ht]
    \centering
    \includegraphics[width=8cm]{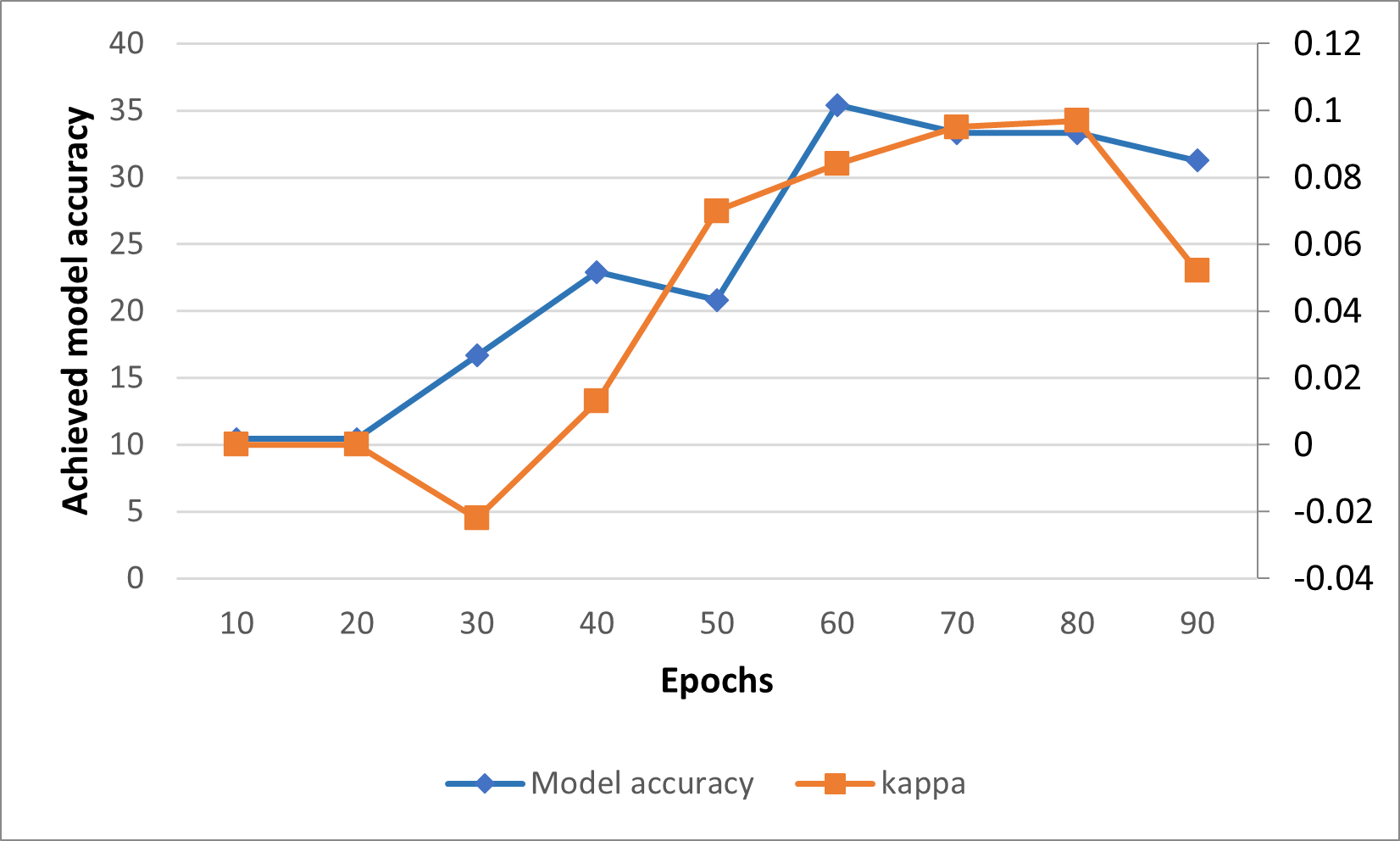}
    \caption{The achieved accuracy after N training epochs and the kappa score between automatically labeled results and feedback.}
    \label{fig:labeling framwork no round accuracy}
\end{figure}

In the first set, we remove the feedback-training process to train on simple transfer learning based on pre-trained imagenet weights. Through the experiment, we extended the training epochs to see whether the result get better through more learning iterations. After that, the trained model was used to predict the transition of randomly selected 100 comic panel pairs. After the prediction, our annotators were asked to label the predicted pairs as well. Therefore we can evaluate the reliability of automatically labeled transitions. The results were in table \ref{fig:labeling framwork no round accuracy} and figure \ref{fig:labeling framwork no round accuracy}. The reliability we used in here was also Cohen's kappa score.

\begin{figure}[ht]
    \centering
    \includegraphics[width=8cm]{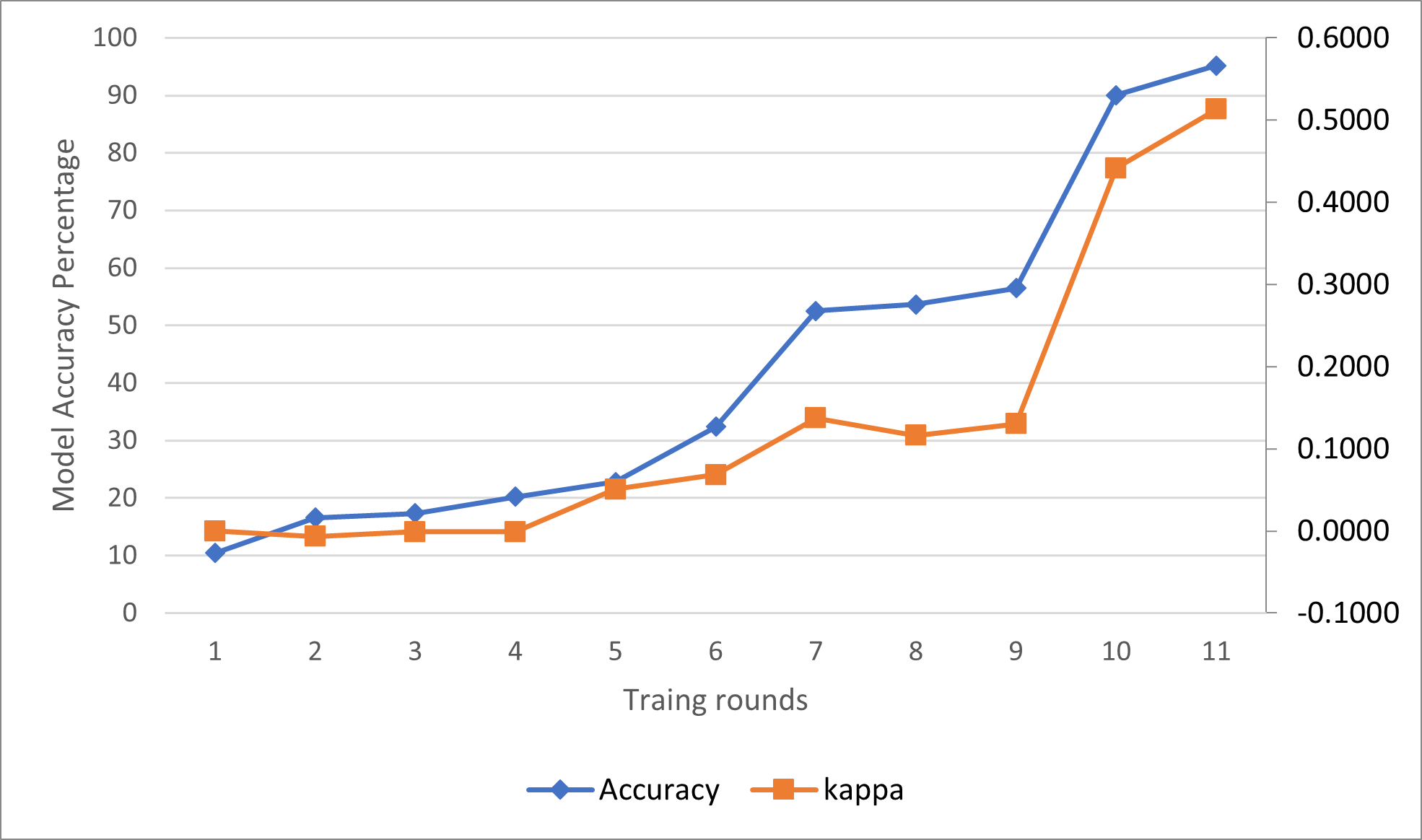}
    \caption{The achieved accuracy after N training rounds, and each round has 10 epochs. And the kappa scores between prediction and feedback.}
    \label{fig:labeling framwork with rounds accuracy}
\end{figure}

For the second set, the experiment was conducted on the labeling framework described above. We had started from our labeled ground truth, then after the training iterations ended. The model tested on a holdout set, and it was asked to predict 100 randomly selected pairs from unlabeled panels. We then collected feedback as in the evaluation step of set 1.  In the next round, according to the feedback, the predicted pairs become two groups. The correct group enlarged the labeled pool, and the combination of holdout set and training set changes based on  0.9, 0.1 proportion. Then the next round of training started. To observe the performance changing after similar iteration numbers, comparing with set 1, we set 10 epochs for each training round.  The results were in table \ref{mixed_books_no_rounds} and figure \ref{fig:labeling framwork with rounds accuracy} provided a clearer view to interpret the result.

The settings shared in both sets were below: First, because it is a classification problem that uses the 6 types of transitions as categories, We evaluate the loss through categorical cross-entropy in the Keras framework, which computes the loss between the true labels and predictions. The optimizer for the model is the RMSprop algorithm, increasing our learning rate to converging faster. We had two regular deeply connected neural network layers after describing the input with the pre-trained VGG16 image-net weights; the top one transferred input into 256 units defined by the relu activation function. We then make the output fall into units with the same number of transition categories with the sigmoid activation function in the second layer.

\begin{table*}
\centering
\begin{tabular}{|p{2.5cm}|p{1cm}|p{1cm}|p{1cm}|p{1cm}|p{1cm}|p{1cm}|p{1cm}|p{1cm}|p{1cm}|}
    \hline
     labeling framework & 10 epochs & 20 epochs & 30 epochs & 40 epochs & 50 epochs & 60 epochs & 70 epochs & 80 epochs & 90 epochs \\
    \hline
    Learning Accu & 10.42\% & 10.42\% & 16.67\% & 22.92\% & 20.83\% & 35.42\% & 33.33\% & 33.33\% & 31.25\%  \\ 
    \hline 
    Cohen's kappa &  0.0 & 0.0 & -0.022& 0.013& 0.07 & 0.084 & 0.095 & 0.097 & 0.052 \\ 
    \hline 

\end{tabular}
    \caption{
    \label{mixed_books_no_rounds} \small{The accuracy after different training rounds and the Cohen's kappa statistic between prediction and feedback.}
    }
\end{table*}


\begin{table*}
\centering
\begin{tabular}{|p{2.5cm}|p{1cm}|p{1cm}|p{1cm}|p{1cm}|p{1cm}|p{1cm}|p{1cm}|p{1cm}|p{1cm}|p{1cm}|p{1cm}|}
    \hline
     labeling framework & 1 round & 2 round & 3 round & 4 round & 5 round & 6 round & 7 round & 8 round & 9 round & 10 round & 11 round  \\
    \hline
    Learning Accu & 10.41\% & 16.56\% & 17.32\% & 20.23\% & 22.81\% & 32.36\% & 52.53\% & 53.67\% & 56.49\% & 90.01\% & 95.19\% \\ 
    \hline 
    Cohen's kappa &  0.0000 & -0.0066 & -0.0012& -0.0012& 0.0506 & 0.0685 & 0.1371 & 0.1163 & 0.1303 & 0.4415 & 0.5131\\ 
    \hline 

\end{tabular}
    \caption{
    \label{mixed_books_with_rounds} \small{The achieved accuracy after N training rounds, and each round has 10 epochs. And the kappa scores between prediction and feedback.}
    }
\end{table*}

Comparing the table \ref{mixed_books_no_rounds} and table \ref{mixed_books_with_rounds}, it is clear that without the feedback-training process, the achieved accuracy is much lower even after many iterations.  Also, the reliability between prediction and feedback is always very low, indicating no agreement. However, the reliability of predictions for the model trained on the feedback-training process reached a moderate level in the end.

\subsection{Transition analysis and clustering}

After the model labeled transitions for the dataset, we then analyzed the transitions in two scopes. The first one was to summarize the transition uses in books. This helped us have a rough view about whether the way authors organize the narrative is different in various stories since the transition implied how separate fragments connected in the story and how readers were guided when reading. Moreover, we retrieve the narrative sequence formed by transitions and calculate the most frequent ones. This showed a characteristic of how authors tend to use transitions. The results were then compared with the real genre labels from the manga109 dataset.

The experiment of transition summarizing had two parts. The first part was to present the transition distribution as vectors and then perform a clustering algorithm to see whether there were any similarities between transition uses. If so, we then curious about if the clustering overlaps with the book genres? Therefore, the second part of the experiment is to group books that share similar genres and compare the intersection between clustering and genre groups. For this experiment, 30 books in manga109 dataset, 22197 panel pairs, were fully labeled and analyzed. 

\begin{figure}
\centering
    \begin{subfigure}{.25\textwidth}
        \centering
        \includegraphics[width=\linewidth]{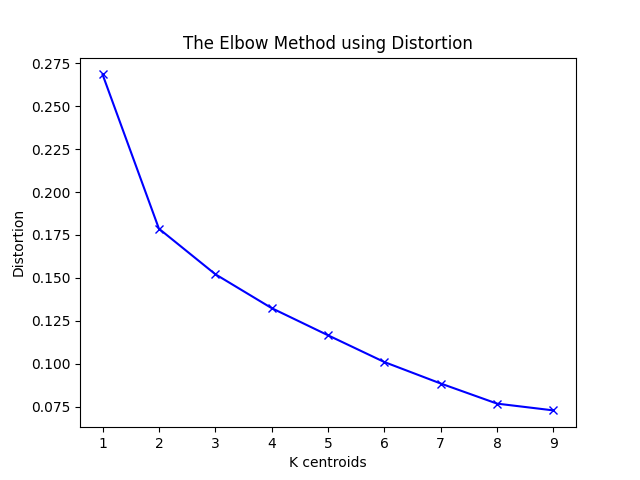}
        \caption{Distortion}
        \label{fig:optimize_k_distortion}
    \end{subfigure}%
    \begin{subfigure}{.25\textwidth}
        \centering
        \includegraphics[width=\linewidth]{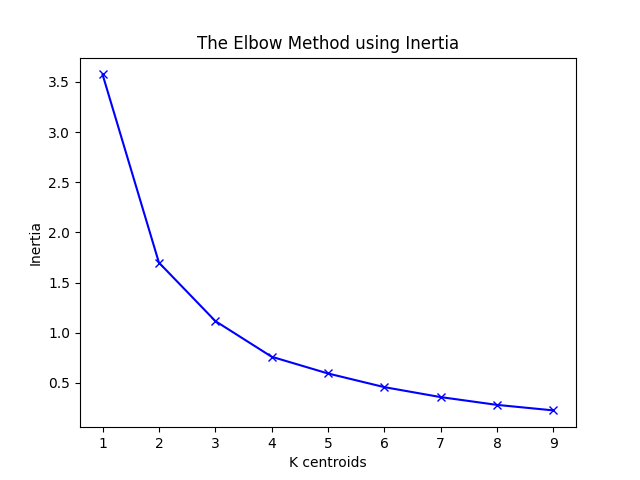}
        \caption{Inertia}
        \label{fig:optimize_k_inertia}
    \end{subfigure}
    \caption{Elbow method using distortion and inertia to decide the feasible number is 4 centers, because the changes become smoother after 4.}
    \label{fig:elbow_method}
\end{figure}

The clustering method here was K-means, and the number of centroids was decided by be elbow method. The figure \ref{fig:elbow_method} showed the distortion and inertia of n centers. We then decide 4 was the feasible number for our data. 

\begin{figure}
\centering
    \begin{subfigure}{.25\textwidth}
        \centering
        \includegraphics[width=\linewidth]{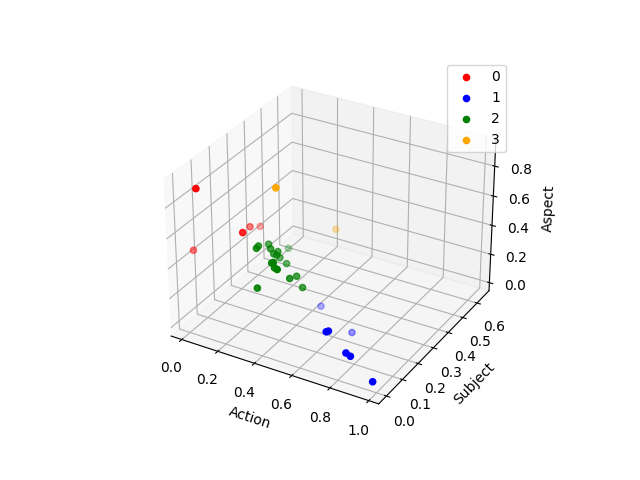}
        \caption{3D view of clustering results, using Action-to-Action, Aspect-to-Aspect, Subject-to-Subject as axis.}
        \label{fig:4_clusters}
    \end{subfigure}%
    \begin{subfigure}{.25\textwidth}
        \centering
        \includegraphics[width=\linewidth]{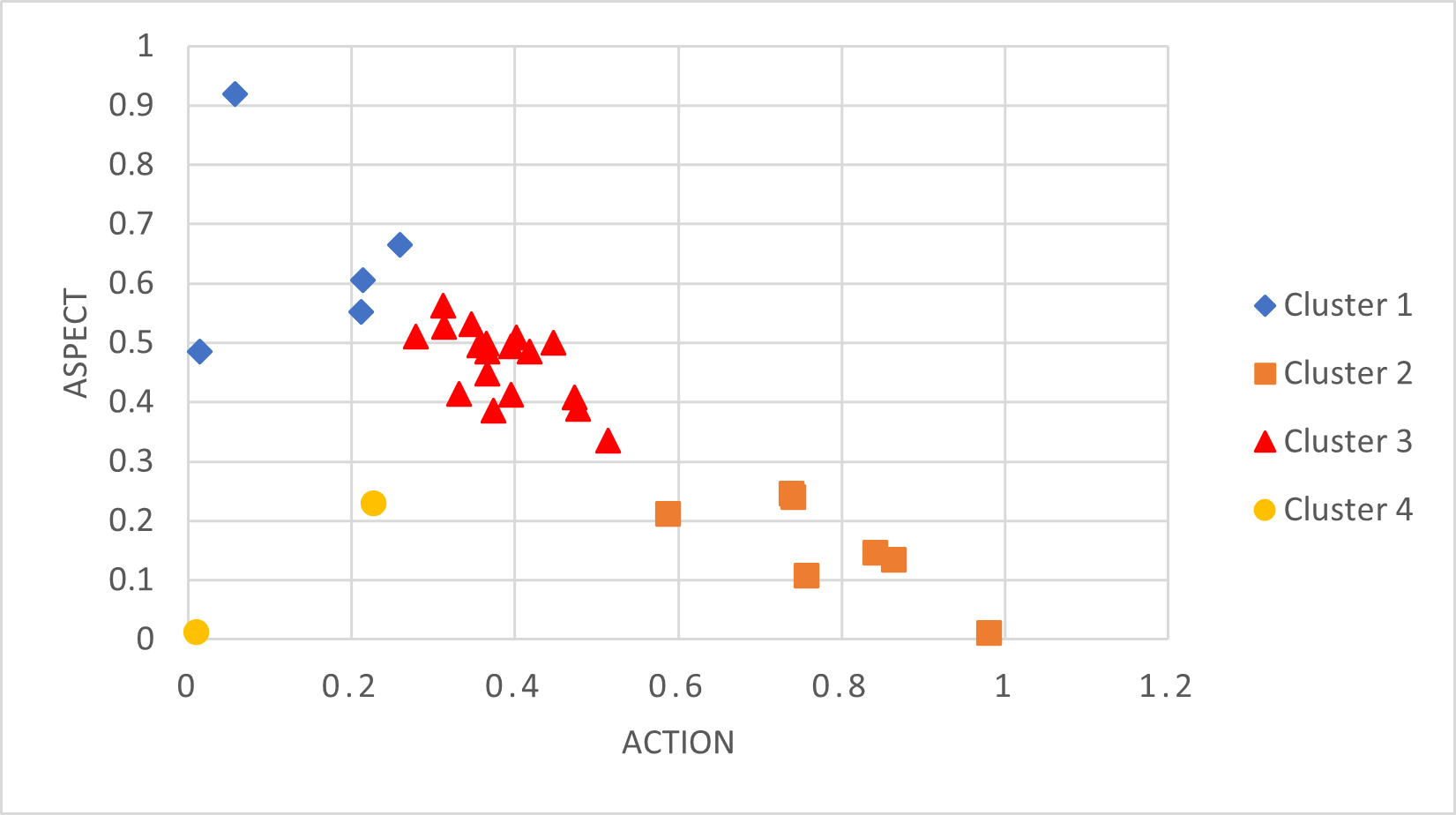}
        \caption{2D view of clustering results, using Action-to-Action, Aspect-to-Aspect as axis.}
        \label{fig:action_aspect_cluster}
    \end{subfigure}
    \begin{subfigure}{.25\textwidth}
        \centering
        \includegraphics[width=\linewidth]{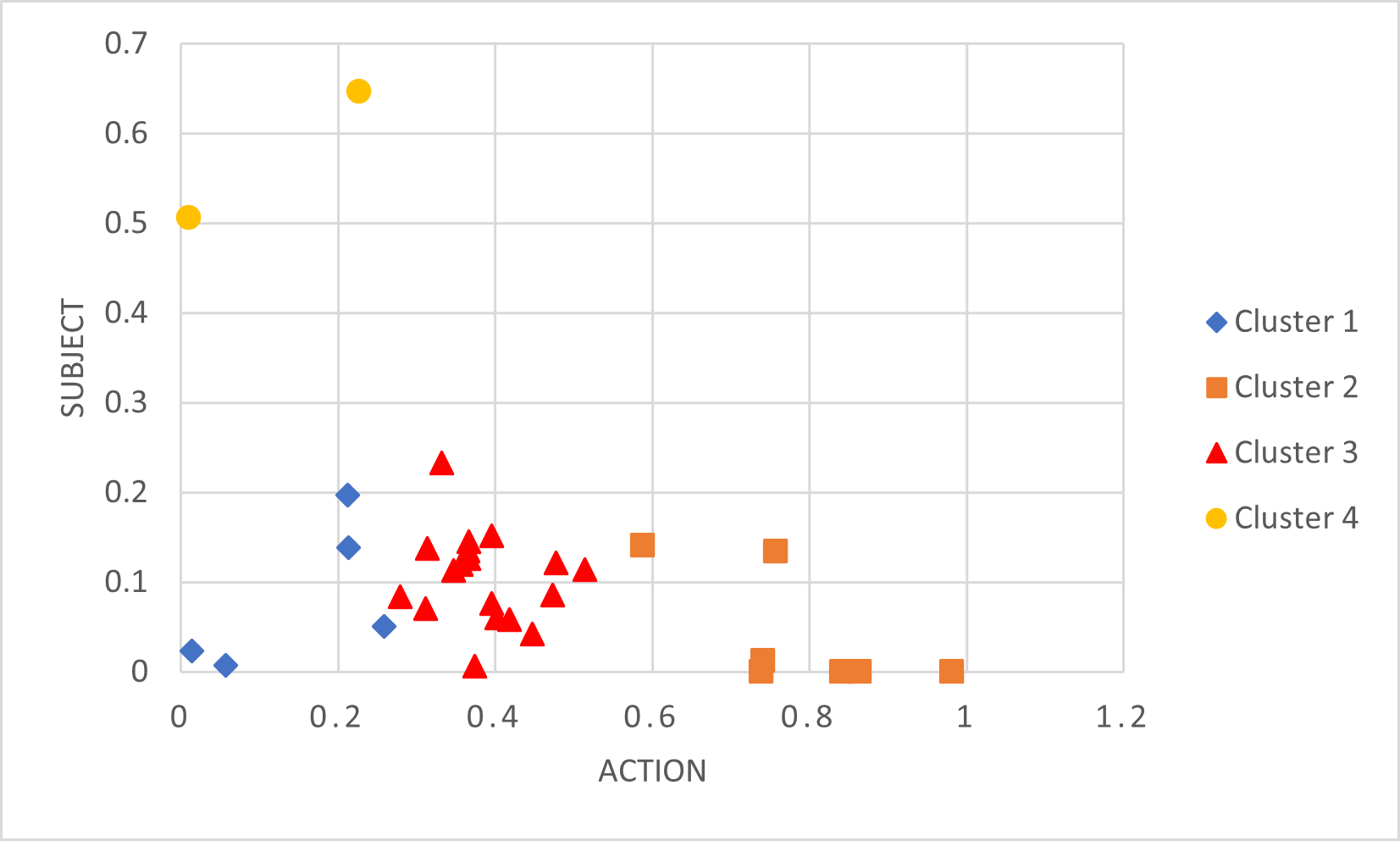}
        \caption{2D view of clustering results, using Action-to-Action, Subject-to-Subject as axis.}
        \label{fig:action_subject_cluster}
    \end{subfigure}%
    \begin{subfigure}{.25\textwidth}
        \centering
        \includegraphics[width=\linewidth]{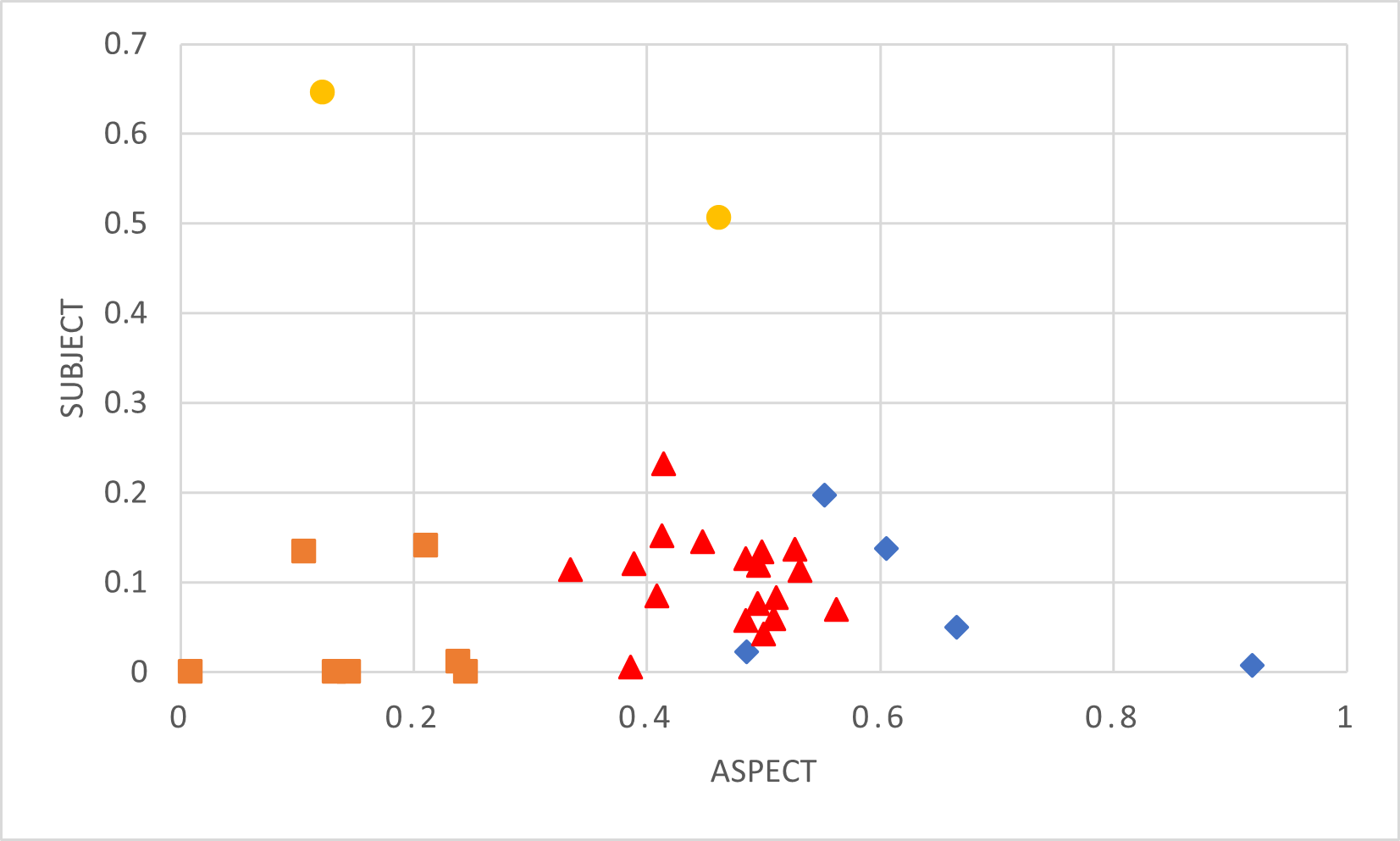}
        \caption{2D view of clustering results, using Aspect-to-Aspect, Subject-to-Subject as axis.}
        \label{fig:subject_aspect_cluster}
    \end{subfigure}
    \caption{The clusering results with 4 centroids.}
    \label{fig:clustering_result}    
\end{figure}

For the labeled book, we found that three types of transition were most significant. The figure \ref{fig:clustering_result} represents the clustering result with 4 centroids using the three types of transitions as the axis. The feature vectors formed by transition use,  we used to represent books and did the clustering, were normalized. 


While Manga109 provided 12 genres, in our labeled books, some genres only have few books; if the amount of books is not sufficient, it is hard to observe the genre overlapping. We then combine the genres into five reasonable groups based on the common characteristic the genres have. The groups are "Romantic," which consists of Love-romance and romantic-comedy." Fiction," composed of Science-fiction and Fantasy. "Action," composed of Battle and Sports. "Plot," by Historical-drama, Suspense, Animals. And finally, remain the Four-panel cartoons unchanged because its' style makes the category non-combinable with other types of comics.

\begin{figure}[ht]
    \centering
    \includegraphics[width=8cm]{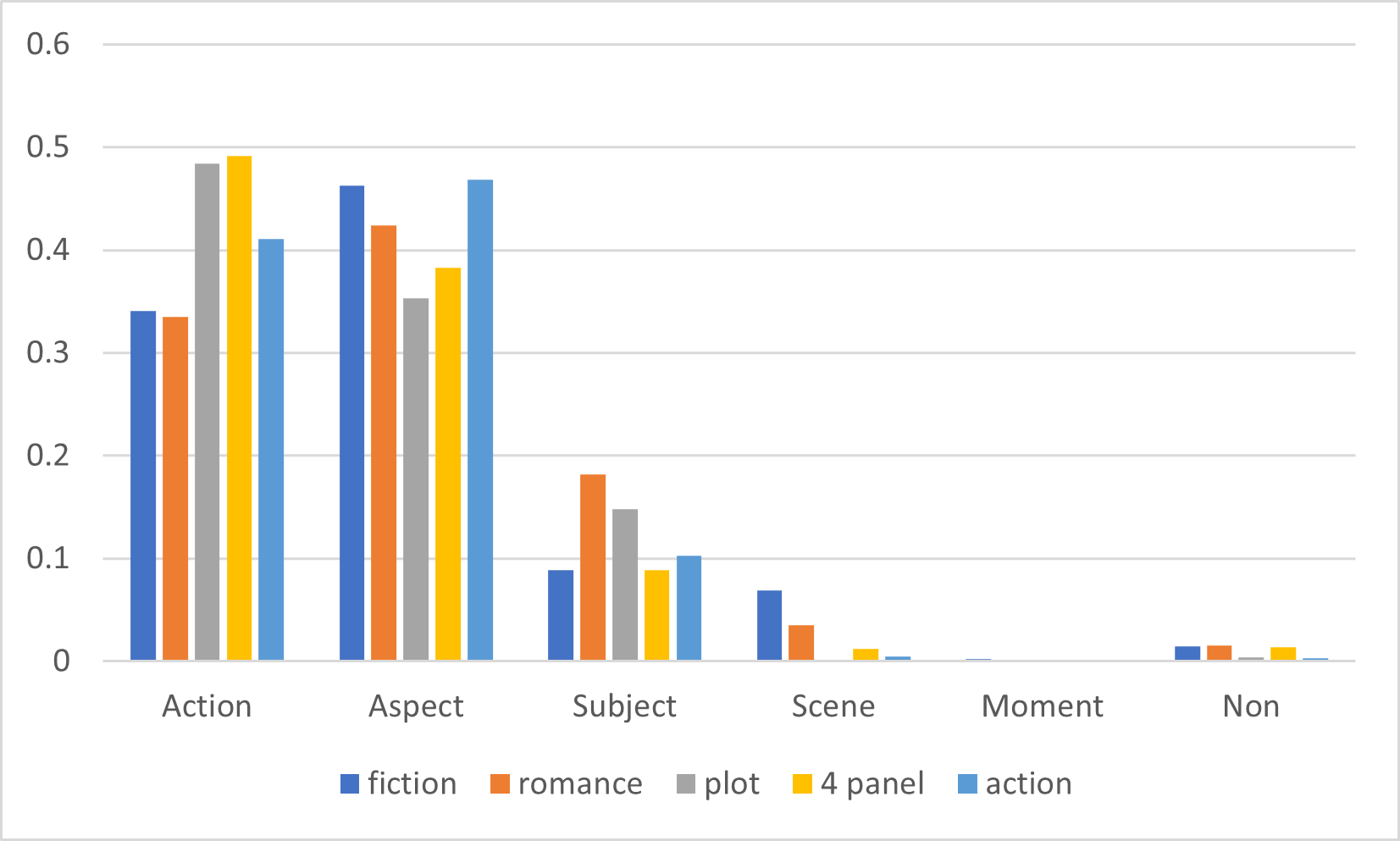}
    \caption{The transition distributions on genres.The number represent the average of normalized distribution.}
    \label{fig:transition_distribution }
\end{figure}

The figure \ref{fig:transition_distribution } displayed the comparison between transition statistics of various groups of genres. Given that the number of transitions in each book is inconsistent, the matrix that employed six types of transitions as the axis was normalized before comparison. 
From the figure, we could observe that more Action-to-Action transitions were involved in storytelling in "Action" "Plot" types, while Aspect-to-aspect transitions seemed to show up more frequently when the story type is  "Fiction" or "Romance." Moreover, the "Romance" stories employed more Subject-to-Subject transitions to connect panels.

The intersection between genre groups and clusters was in the table \ref{intersection_groups} to know whether the genre correlates with the use transitions. From the results, we could see that while the stories intersected with multiple clusters, each genre type of story seemed to focus more on some pattern of transition uses rather than evenly scatter in every cluster.  For example, the "Romance." and "Fiction." groups overlap more with cluster 3, which has more Aspect-to-Aspect transition than other types. This also matches the bar chart figure) which summarized the overall use of transitions.

\begin{table}
\centering
\begin{tabular}{|p{3cm}|p{0.8cm}|p{0.8cm}|p{0.8cm}|p{0.8cm}|}
    \hline
     Genres & clus1 & clus2 & clus3 & clus4\\
    \hline 
    Action(Battle/Sports) & 0.40 & 0.20 & 0.40& 0.00\\
    \hline 
    Romance(Love romance/Romantic comedy) & 0.14&  0.00& 0.71& 0.14\\
    \hline 
    4 panels cartoon &  0.00& 0.25 & 0.75 & 0.00\\ 
    \hline     
    fiction(Fantasy/Science fiction) & 0.11& 0.33& 0.67& 0.00 \\
    \hline  
    plot(Historical drama/Animal/Suspense) & 0.00& 0.50& 0.50& 0.00\\
    \hline     
\end{tabular}
    \caption{
    \label{intersection_groups} \small{Intersections between clusters based on transition distribution and real genres}
    }
\end{table}

\subsubsection{Transition sequences analysis}

The last analysis we would like to provide from the generated transition labels was the transition sequences used most frequently in different stories. From various genre groups of books, we collected different lengths of transition sequences. The table \ref{transition_sequences} contained the results.The statistic results said that the sequences looks have many repetitions in them. For example, when an Aspect-to-Aspect transition showed up, it is highly possible, followed by another Aspect-to-Aspect transition. A similar phenomenon also applies to other types of transitions; we assess that maybe whenever an event is introduced in a story, it won't end immediately. The examples we had in mind are fighting scenes or conversation scenes, making the readers' focus switching between characters or view angles. Moreover,  since the transitions bridge the inter-panel gaps, their repetition may imply the rhythm of storytelling. As shown in the table \ref{transition_sequences}, starting from sequences of length 3, the repeated phenomenon ends, and different types of transitions begin to mix in. Further analysis is a possibility that helps to studies the discourse of comic stories.

\begin{table*}
\centering
\begin{tabular}{|p{4cm}|p{2.2cm}|p{2.2cm}|p{2.2cm}|p{2.2cm}|p{2.2cm}|}
    \hline
     Categories & Action & Romance & 4 panels cartoon & Fiction & Plot\\
    \hline
    Top 1 sequence, with length 1 & [ASP]& [ASP]& [ACT]& [ACT]& [ACT]\\ 
    \hline 
    Top 2 sequence, with length 1 & [ACT]& [ACT]& [ASP]& [ASP] & [ASP]\\ 
    \hline 
    Top 3 sequence, with length 1 & [SUB] & [SUB] & [SUB] & [SUB] & [SUB]\\ 
    \hline     
    Top 4 sequence, with length 1 & [SCE] & [SCE] & [NON] & [SCE] & [NON]\\ 
    \hline     

    \hline
    Top 1 sequence, with length 2 & [ASP, ASP]& [ASP, ASP]& [ACT,ACT]& [ACT,ACT] & [ACT,ACT] \\ 
    \hline 
    Top 2 sequence, with length 2 & [ACT, ACT]& [ACT, ACT]& [ASP, ASP]& [ASP, ASP]& [ASP, ASP]\\ 
    \hline 
    Top 3 sequence, with length 2 & [SUB, SUB]& [SUB, SUB]& [SUB, SUB]& [SUB, SUB]& [SUB, SUB]\\ 
    \hline     
    Top 4 sequence, with length 2 & [ACT,ASP]& [ASP, ACT]& [ASP, ACT]& [ASP, ACT]& [ASP, ACT]\\ 
    \hline     
    \hline   
    Top 1 sequence, with length 3 & [ASP, ASP, ASP]& [ASP, ASP, ASP]&[ACT, ACT,ACT] & [ACT, ACT,ACT]& [ACT, ACT,ACT]\\ 
    \hline 
    Top 2 sequence, with length 3 &[ACT, ACT,ACT] & [SUB, SUB, SUB]&[ASP, ASP, ASP]& [ASP, ASP, ASP]& [ASP, ASP, ASP]\\ 
    \hline 
    Top 3 sequence, with length 3 &[SUB, SUB, SUB] & [ACT, ACT,ACT]& [SUB, SUB, SUB]& [SUB, SUB, SUB]& [SUB, SUB, SUB]\\ 
    \hline     
    
    \multirow{4}{*}{Top 4 sequence, with length 3}
    &[ACT,ASP,ASP] & [ASP,ASP,ACT] & [ASP,ACT,ACT] & [ASP,ASP,ACT] & [ASP,ACT,ACT]\\
    &[ASP,ASP,ACT] & [ACT,ASP,ASP] & [ACT,ASP,ASP] & [ACT,ACT,ASP] & [ACT,ACT,ASP]\\
    &[ACT,ACT,ASP]&  [ASP,ACT,ACT]&  [ACT,ACT,ASP]& & \\
    & &[ACT,ACT,ASP]&[ASP,ASP,ACT] & & \\
    \hline     
    \hline   
    Top 1 sequence, with length 4 & [AS,AS,AS,AS]& [AS,AS,AS,AS]& [AC,AC,AC,AC]& [AC,AC,AC,AC]& [AC, AC,AC,AC]\\ 
    \hline 
    Top 2 sequence, with length 4 & [AC,AC,AC,AC]&[AC,AC,AC,AC] & [AS,AS,AS,AS]& [AS,AS,AS,AS]& [AS,AS,AS,AS]\\ 
    \hline 
    Top 3 sequence, with length 4 & [SU,SU,SU,SU]& [SU,SU,SU,SU] & [SU,SU,SU,SU]&[SU,SU, SU,SU]&[SU,SU,SU,SU] \\ 
    \hline 
    \multirow{3}{*}{Top 4 sequence, with length 4}
    & [AS,AS,AS,AC] & [AC,AC,AC,AS] &  [AC, AC,AC,AS] & [AS,AC,AC,AC] &[AC,AC,AC,AS]\\
    & & [AC,AS,AS,AS] &  & & \\
    &&[AC,AC,AS,AS] & & & \\
    & && & & \\

    \hline

\end{tabular}
    \caption{
    \label{transition_sequences} \small{The most frequent transition sequences used in different genres.The most frequent transition sequences used in different genres.Action-to-Action(ACT, AC), Aspect-to-Aspect(ASP, AS), Subject-to-Subject(SUB, SU), Scene-to-Scene(SCE,SC), Moment-to-Moment(MOM, MO), Non-sequitur(NON, NO).}
    }
\end{table*}





\section{conclusion and future work}
In this paper, we began by motivating the use of manga-style visual comics as an interesting domain for the study of narrative semantics in media. We started with a principled application of panel transition labels to automatically annotate the dataset from a model trained with input from human labels. We systematically analyzed the labeling results by clustering to observe the similarity that transition uses might have. And conducted a further comparison of results and real comic genres to show overlaps. This suggests the possible relationship between comic genres and panel transition features in addition to low-level compositional features extracted through computer vision. our results are promising and we hope by sharing the dataset, annotation methods, models, and the overall methodology with the community we will inspire productive work on media analysis. We will re-state the contributions of this paper in three aspects. First, a principled connection between practitioner specified best practices and semantic labelling. Second, a dataset and a workflow for annotations and analysis, Finally, a specific demonstration of detailed analysis of genre classification in manga as well as sequential patterns of panel transitions for analyzing the richness of expression in this medium.

\vfill\null

\bibliographystyle{natdin}
\bibliography{reference}

\begin{thebibliography}{22}


\providecommand{\natexlab}[1]{#1}
\providecommand{\url}[1]{\texttt{#1}}
\makeatletter
\newcommand{\dinatlabel}[1]%
{\ifNAT@numbers\else\NAT@biblabelnum{#1}\fi}
\makeatother
\expandafter\ifx\csname urlstyle\endcsname\relax
  \providecommand{\doi}[1]{doi: #1}\else
  \providecommand{\doi}{doi: \begingroup \urlstyle{rm}\Url}\fi

\bibitem[Choro{\'s}(2018)]{choros2018video}
\dinatlabel{Choro{\'s} 2018} \textsc{Choro{\'s}}, Kazimierz:
\newblock Video genre classification based on length analysis of temporally aggregated video shots.
\newblock {In: }\emph{International Conference on Computational Collective Intelligence} Springer, 2018, S. 509--518

\bibitem[Cohen(1960)]{cohen1960coefficient}
\dinatlabel{Cohen 1960} \textsc{Cohen}, Jacob:
\newblock A coefficient of agreement for nominal scales.
\newblock {In: }\emph{Educational and psychological measurement} 20 (1960), Nr. 1, S. 37--46

\bibitem[Cohn(2019)]{cohn2019pins}
\dinatlabel{Cohn 2019} \textsc{Cohn}, N:
\newblock Your Brain on Comics: A Cognitive Model of Visual Narrative Comprehension.
\newblock {In: }\emph{Topics in cognitive science}  (2019)

\bibitem[Daiku u.\,a.(2017)Daiku, Augereau, Iwata, u. Kise]{daiku2017comic}
\dinatlabel{Daiku u.\,a. 2017} \textsc{Daiku}, Yuki ; \textsc{Augereau}, Olivier ; \textsc{Iwata}, Motoi  ; \textsc{Kise}, Koichi:
\newblock Comic story analysis based on genre classification.
\newblock {In: }\emph{2017 14th IAPR International Conference on Document Analysis and Recognition (ICDAR)} Bd.~3 IEEE, 2017, S. 60--65

\bibitem[Daiku u.\,a.(2018)Daiku, Iwata, Augereau, u. Kise]{daiku2018comics}
\dinatlabel{Daiku u.\,a. 2018} \textsc{Daiku}, Yuki ; \textsc{Iwata}, Motoi ; \textsc{Augereau}, Olivier  ; \textsc{Kise}, Koichi:
\newblock Comics story representation system based on genre.
\newblock {In: }\emph{2018 13th IAPR International Workshop on Document Analysis Systems (DAS)} IEEE, 2018, S. 257--262

\bibitem[Doshi u. Zadrozny(2018)]{doshi2018movie}
\dinatlabel{Doshi u. Zadrozny 2018} \textsc{Doshi}, Pratik ; \textsc{Zadrozny}, Wlodek:
\newblock Movie genre detection using topological data analysis.
\newblock {In: }\emph{International Conference on Statistical Language and Speech Processing} Springer, 2018, S. 117--128

\bibitem[Iyyer u.\,a.(2017)Iyyer, Manjunatha, Guha, Vyas, Boyd-Graber, Daume, u. Davis]{iyyer2017amazing}
\dinatlabel{Iyyer u.\,a. 2017} \textsc{Iyyer}, Mohit ; \textsc{Manjunatha}, Varun ; \textsc{Guha}, Anupam ; \textsc{Vyas}, Yogarshi ; \textsc{Boyd-Graber}, Jordan ; \textsc{Daume}, Hal  ; \textsc{Davis}, Larry~S.:
\newblock The amazing mysteries of the gutter: Drawing inferences between panels in comic book narratives.
\newblock {In: }\emph{Proceedings of the IEEE Conference on Computer Vision and Pattern Recognition}, 2017, S. 7186--7195

\bibitem[Kundalia u.\,a.(2020)Kundalia, Patel, u. Shah]{kundalia2020multi}
\dinatlabel{Kundalia u.\,a. 2020} \textsc{Kundalia}, Kaushil ; \textsc{Patel}, Yash  ; \textsc{Shah}, Manan:
\newblock Multi-label movie genre detection from a movie poster using knowledge transfer learning.
\newblock {In: }\emph{Augmented Human Research} 5 (2020), Nr. 1, S. 1--9

\bibitem[Martens u.\,a.(2020)Martens, EDU, Cardona-Rivera, u. EDU]{martens2020visual}
\dinatlabel{Martens u.\,a. 2020} \textsc{Martens}, Chris ; \textsc{EDU}, NCSU ; \textsc{Cardona-Rivera}, Rogelio~E.  ; \textsc{EDU}, UTAH:
\newblock The Visual Narrative Engine: A Computational Model of the Visual Narrative Parallel Architecture.
\newblock {In: }\emph{8th Annual Conference on Advances in Cognitive Systems}, 2020

\bibitem[McCloud(2006)]{mccloud2006making}
\dinatlabel{McCloud 2006} \textsc{McCloud}, Scott:
\newblock \emph{Making comics: Storytelling secrets of comics, manga and graphic novels}.
\newblock Harper New York, 2006

\bibitem[McCloud u. Manning(1998)]{mccloud1998understanding}
\dinatlabel{McCloud u. Manning 1998} \textsc{McCloud}, Scott ; \textsc{Manning}, AD:
\newblock Understanding comics: The invisible art.
\newblock {In: }\emph{IEEE Transactions on Professional Communications} 41 (1998), Nr. 1, S. 66--69

\bibitem[McCloud u. Martin(1993)]{mccloud1993understanding}
\dinatlabel{McCloud u. Martin 1993} \textsc{McCloud}, Scott ; \textsc{Martin}, Mark:
\newblock \emph{Understanding comics: The invisible art}. Bd. 106.
\newblock Kitchen sink press Northampton, MA, 1993

\bibitem[McHugh(2012)]{mchugh2012interrater}
\dinatlabel{McHugh 2012} \textsc{McHugh}, Mary~L.:
\newblock Interrater reliability: the kappa statistic.
\newblock {In: }\emph{Biochemia medica} 22 (2012), Nr. 3, S. 276--282

\bibitem[Park u. Matsushita(2019)]{park2019estimating}
\dinatlabel{Park u. Matsushita 2019} \textsc{Park}, Byeongseon ; \textsc{Matsushita}, Mitsunori:
\newblock Estimating comic content from the book cover information using fine-tuned VGG model for comic search.
\newblock {In: }\emph{International Conference on Multimedia Modeling} Springer, 2019, S. 650--661

\bibitem[Pratt(2009)]{pratt2009narrative}
\dinatlabel{Pratt 2009} \textsc{Pratt}, Henry~J.:
\newblock Narrative in comics.
\newblock {In: }\emph{The Journal of Aesthetics and Art Criticism} 67 (2009), Nr. 1, S. 107--117

\bibitem[Pritsos u.\,a.(2019)Pritsos, Rocha, u. Stamatatos]{pritsos2019open}
\dinatlabel{Pritsos u.\,a. 2019} \textsc{Pritsos}, Dimitrios ; \textsc{Rocha}, Anderson  ; \textsc{Stamatatos}, Efstathios:
\newblock Open-Set Web Genre Identification Using Distributional Features and Nearest Neighbors Distance Ratio.
\newblock {In: }\emph{European Conference on Information Retrieval} Springer, 2019, S. 3--11

\bibitem[Shambharkar u.\,a.(2020)Shambharkar, Anand, u. Kumar]{shambharkar2020survey}
\dinatlabel{Shambharkar u.\,a. 2020} \textsc{Shambharkar}, Prashant~G. ; \textsc{Anand}, Anshul  ; \textsc{Kumar}, Anshul:
\newblock A Survey Paper on Movie Trailer Genre Detection.
\newblock {In: }\emph{2020 International Conference on Computing and Data Science (CDS)} IEEE, 2020, S. 238--244

\bibitem[Simeone u.\,a.(2017)Simeone, Santos-Rodr{\'\i}guez, McVicar, Lijffijt, u. De~Bie]{simeone2017hierarchical}
\dinatlabel{Simeone u.\,a. 2017} \textsc{Simeone}, Paolo ; \textsc{Santos-Rodr{\'\i}guez}, Ra{\'u}l ; \textsc{McVicar}, Matt ; \textsc{Lijffijt}, Jefrey  ; \textsc{De~Bie}, Tijl:
\newblock Hierarchical novelty detection.
\newblock {In: }\emph{International Symposium on Intelligent Data Analysis} Springer, 2017, S. 310--321

\bibitem[Sreeja u. Kovoor(2019)]{sreeja2019towards}
\dinatlabel{Sreeja u. Kovoor 2019} \textsc{Sreeja}, MU ; \textsc{Kovoor}, Binsu~C.:
\newblock Towards genre-specific frameworks for video summarisation: A survey.
\newblock {In: }\emph{Journal of Visual Communication and Image Representation} 62 (2019), S. 340--358

\bibitem[Tafreshi u. Diab(2018)]{tafreshi2018emotion}
\dinatlabel{Tafreshi u. Diab 2018} \textsc{Tafreshi}, Shabnam ; \textsc{Diab}, Mona:
\newblock Emotion detection and classification in a multigenre corpus with joint multi-task deep learning.
\newblock {In: }\emph{Proceedings of the 27th international conference on computational linguistics}, 2018, S. 2905--2913

\bibitem[Yadav u. Vishwakarma(2020)]{yadav2020unified}
\dinatlabel{Yadav u. Vishwakarma 2020} \textsc{Yadav}, Ashima ; \textsc{Vishwakarma}, Dinesh~K.:
\newblock A unified framework of deep networks for genre classification using movie trailer.
\newblock {In: }\emph{Applied Soft Computing} 96 (2020), S. 106624

\bibitem[Yu u.\,a.(2020)Yu, Lu, Li, u. Liu]{yu2020asts}
\dinatlabel{Yu u.\,a. 2020} \textsc{Yu}, Yitong ; \textsc{Lu}, Ziyu ; \textsc{Li}, Yang  ; \textsc{Liu}, Delong:
\newblock ASTS: attention based spatio-temporal sequential framework for movie trailer genre classification.
\newblock {In: }\emph{Multimedia Tools and Applications}  (2020), S. 1--16

\end{thebibliography}

\end{document}